\documentclass{article}

\usepackage{microtype}
\usepackage{graphicx}
\usepackage{subfigure}
\usepackage{booktabs} 

\usepackage{hyperref}

\usepackage[accepted]{icml2024/icml2024}

\usepackage{amsmath}
\usepackage{amssymb}
\usepackage{mathtools}
\usepackage{amsthm}
\usepackage{bbm}

\usepackage{dblfloatfix}
\usepackage{xcolor}
\usepackage{tikz}
\definecolor{redd}{HTML}{b30326}
\definecolor{bloo}{HTML}{3a4cc0}

\newcommand*\rdot{\tikz[baseline=(char.base)]{
    \node[shape=circle,fill=redd,text=redd,minimum size=5pt,inner sep=0pt] (char) {x};}}
\newcommand*\bdot{\tikz[baseline=(char.base)]{
    \node[shape=circle,fill=bloo,text=bloo,minimum size=5pt,inner sep=0pt] (char) {x};}}
\definecolor{ours}{HTML}{219675}
\newcommand{\ours}{\textcolor{ours}{\textbf{T-CREx}}}
\definecolor{ourstaulo}{HTML}{1f77b4}
\newcommand{\ourstaulo}{\textcolor{ourstaulo}{\textbf{T-CREx\textsubscript{0.9}}}}
\definecolor{ourstauhi}{HTML}{2ca02c}
\newcommand{\ourstauhi}{\textcolor{ourstauhi}{\textbf{T-CREx\textsubscript{0.99}}}}
\definecolor{ares}{HTML}{ff7f0e}
\newcommand{\ares}{\textcolor{ares}{\textbf{AReS}}}
\definecolor{lore}{HTML}{d62728}
\newcommand{\lore}{\textcolor{lore}{\textbf{LORE}}}
\definecolor{rfocse}{HTML}{9467bd}
\newcommand{\rfocse}{\textcolor{rfocse}{\textbf{RF-OCSE}}}

\definecolor{circlefill}{HTML}{bce8da}
\newcommand*\annot[1]{\tikz[baseline=(char.base)]{
\node[shape=circle,draw=ours,fill=circlefill,minimum size=13pt,inner sep=0pt] (char) {#1};}}

\usepackage[capitalize,noabbrev]{cleveref}

\theoremstyle{plain}

\theoremstyle{definition}

\theoremstyle{remark}

\usepackage[textsize=tiny]{todonotes}

\icmltitlerunning{Counterfactual Metarules for Local and Global Recourse}

\begin{document}

\twocolumn[
\icmltitle{\
Counterfactual Metarules for Local and Global Recourse
}



\icmlsetsymbol{equal}{*}

\begin{icmlauthorlist}

\icmlauthor{Tom Bewley}{jpmc}
\icmlauthor{Salim I. Amoukou}{jpmc}
\icmlauthor{Saumitra Mishra}{jpmc}
\icmlauthor{Daniele Magazzeni}{jpmc}
\icmlauthor{Manuela Veloso}{jpmc}

\end{icmlauthorlist}

\icmlaffiliation{jpmc}{J.P. Morgan AI Research}
\icmlcorrespondingauthor{Tom Bewley}{tom.bewley@jpmorgan.com}

\icmlkeywords{explainability, interpretability, XAI, counterfactual explanation, algorithmic recourse, tree-based models, rule-based models}

\vskip 0.3in
]

\printAffiliationsAndNotice{}  

\begin{abstract}

We introduce \ours{}, a novel model-agnostic method for local and global counterfactual explanation (CE), which summarises recourse options for both individuals and groups in the form of human-readable rules. It leverages tree-based surrogate models to learn the
counterfactual rules, alongside \textit{metarules} denoting their regions of optimality, providing both a global analysis of model behaviour and diverse recourse options for users.
Experiments indicate that \ours{} achieves superior aggregate performance over existing rule-based baselines on a range of CE desiderata, while being orders of magnitude faster to run.
\end{abstract}

\vspace{-0.5cm}
\section{Introduction}

\vspace{-0.1cm}
Counterfactual explanation (CE), which describes how input features could be changed to alter a model's output, is a ubiquitous technique in eXplainable AI (XAI).
As AI models make increasingly many decisions that impact human users, CEs provide a foundation for recourse, whereby users act to change an adverse output (e.g. loan rejection) to a desirable one (e.g. acceptance) \cite{wachter2017counterfactual}.

\vspace{-0.1cm}
While the most basic objective is to find one CE example per instance, several works have generalised this to find either a single group-level
CE for a \textit{set} of instances \cite{carrizosa2024mathematical}, or a diverse \textit{set} of CEs for a single instance \cite{mothilal2020explaining}, optionally summarising each kind of set using human-interpretable rules \cite{Kanamori2022counterfactual,rawal2020beyond}. Both generalisations bring benefits: group-level CEs provide a route to globally analysing a model's behaviour and subgroup fairness properties, while diverse CEs present users with a range of recourse options, which increases the chance of one being
practically
actionable. In addition, summarising diverse CEs in a compact rule-based form mitigates the information overload that may result from diversity, and provides robustness to the problem of recourse noise \cite{pawelczyk2022algorithmic} by specifying an extended range of values that each feature can take.

\begin{figure}
    \centering
    \includegraphics[width=\columnwidth]{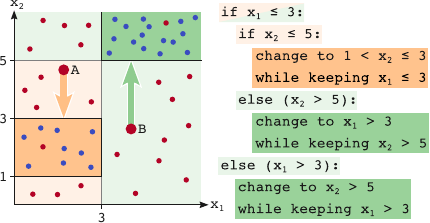}
    \vspace{-0.5cm}
    \caption{
Application of \ours{} to a binary classifier, producing CEs for
the red class
\rdot.
Shown are two rules (dark green/orange) for which $\geq 90\%$ of contained points have
the alternative blue class
\bdot. The rules are paired with metarules denoting regions of the input space where each rule is an optimal CE (light green/orange), which can be interpreted as follows.
For inputs with $x_1\leq 3$ and $x_2\leq 5$, the orange rule is optimal because it requires changing only one feature (sparsity). Elsewhere, the green rule is preferred because it contains a greater number of points (feasibility). The two rules and three metarules can be combined to create a global textual summary of all recourse options (shown on the right), which in turn enables the near-instantaneous generation of a CE for any single instance (e.g. A or B) via a simple lookup.   
    }
    \label{fig:page1}
\vspace{-0.4cm}
\end{figure}

We present \ours{}, a method combining both forms of generalisation. It uses tree-based surrogate models to learn counterfactual rules denoting regions of the input space where a model changes its output with high probability, alongside \textit{metarules} denoting regions
for which each rule is an optimal CE. The optimal rule for each input (and thus for each metarule) is a joint function of the number of features that must change to satisfy the rule (sparsity) and the rule's population under a given data distribution (feasibility).
The method is model-agnostic, requiring only black box access to the target model, and handles both numerical and categorical features and both classification and regression problems.
Figure~\ref{fig:page1} shows the rules and metarules produced by \ours{} for a simple binary classification example.

\vspace{-0.1cm}
We are aware of one prior method that uses rules to provide diverse CEs for groups of instances comparably to our rule/metarule formulation \cite{rawal2020beyond}, and several others that find rule-based CEs for single instances \cite{guidotti2019factual,fernandez2020random}. Through direct comparison, we show that \ours{} reliably matches or outperforms these baselines on various CE desiderata, and is orders of magnitude faster to run.
We also
present a selection of qualitative analyses, aided by our ability to represent rules and metarules in a human-readable tree structure.

\vspace{-0.1cm}
Our contributions can be summarised as follows:
\begin{itemize}
    \vspace{-0.25cm}
    \item A general formulation of the rule/metarule approach to CE, which could in principle be applied to any black box predictive model, and a concrete instantiation for real vector input spaces using hyperrectangles.
    \vspace{-0.15cm}
    \item The \ours{} algorithm, which uses trees to learn hyperrectangular counterfactual rules and metarules for a specific sparsity- and feasibility-based cost function.
    \vspace{-0.15cm}
    \item Experiments demonstrating \ours{}'s strong quantitative performance relative to baseline methods, most notably in terms of computational efficiency.
    \vspace{-0.15cm}
    \item An exploration of the qualitative structure of learnt counterfactual rules and metarules, demonstrating how they provide both local and global insights.
\end{itemize}

\section{Counterfactual Rules and Metarules}

Let $\mathcal{X}$ and $\mathcal{Y}$ be input and output spaces and $f:\mathcal{X}\to\mathcal{Y}$ be a model. Given $x^0\in\mathcal{X}$
with model output $f(x^0)$ and a set of target outputs $Y^*\subseteq \mathcal{Y}\setminus\{f(x^0)\}$, counterfactual point explanation (CPE)  seeks the lowest-cost transformation of $x^0$ that yields an output in $Y^*$. Formally, the aim is to find
\vspace{-0.1cm}
\begin{equation} \label{eq:CPE}
    \text{CPE}(x^0\ |\ Y^*, P_\mathcal{X})=\hspace{-0.2cm}\underset{x^i\in \mathcal{X}:f(x^i)\in{Y^*}}{\text{argmin}} \hspace{-0.2cm} \text{cost}(x^0, x^i\ |\ P_\mathcal{X}),
    \vspace{-0.1cm}
\end{equation}
where the cost may depend only on the relationship between $x^0$ and $x^i$ (e.g. their separation under some distance metric), or also on a given distribution $P_\mathcal{X}$ of \textit{realistic} inputs (e.g. assigning low cost to points in high-density regions, which are seen as feasible recourse targets \cite{poyiadzi2020face}).

\vspace{-0.1cm}
In this work, we seek counterfactuals expressed not as single points, but as \textit{rules} covering regions of the input space where the model $f$ gives an output in $Y^*$ with high probability, thereby summarising a range of possible recourse options.
Formally, we define rules 
$\mathcal{R}\subseteq 2^\mathcal{X}$ as a class of subsets of $\mathcal{X}$.
To ensure that the resultant explanations are understandable to humans, we require these rules to satisfy some definition of interpretability; one such definition is adopted
below.
We say that a rule $R^i\in\mathcal{R}$ is \textit{valid} for a given realistic distribution $P_\mathcal{X}$ and output set $Y^*$ if it contains at least a fraction $\rho \in (0,1]$ of realistic inputs, and $f$ gives an output in $Y^*$ for at least a fraction $\tau\in(0,1]$ of those inputs:\footnote{
In practice, these values are estimated over a finite sample.
}
\vspace{-0.1cm}
\begin{equation} \label{eq:rule_is_valid}
    \text{val}(R^i) = \mathbbm{1}[\text{feasibility}(R^i) {\geq} \rho\land\text{accuracy}(R^i) {\geq} \tau],
    \vspace{-0.1cm}
\end{equation}
where
\begin{equation} \label{eq:feasibility}
    \text{feasibility}(R^i)= \mathbb{P}_{x\sim P_\mathcal{X}}\{x\in R^i\};
\end{equation}
\begin{equation} \label{eq:accuracy}
    \text{accuracy}(R^i) = \mathbb{P}_{x\sim P_\mathcal{X}:x\in R^i}\{f(x)\in Y^*\}.
\end{equation}

Let the set of \textit{maximal}-valid rules be those that are not proper subsets of any other valid rule:
\vspace{-0.1cm}
\begin{equation} \label{eq:maximal_valid_rules}
    \mathcal{R}^{\text{max}}_{Y^*} {=} \{R^i {\in} \mathcal{R}\ |\ \text{val}(R^i) \land (\nexists R^j {\in} \mathcal{R} : R^i {\subset} R^j \land \text{val}(R^j))\}.
    \vspace{-0.1cm}
\end{equation}
In other words, a maximal-valid rule is one that cannot be made any larger without violating the validity conditions.

\vspace{-0.1cm}
Given a set of candidate maximal-valid rules, the aim of counterfactual rule explanation (CRE) is to find the one that minimises some cost function,
\vspace{-0.1cm}
\begin{equation} \label{eq:CRE}
    \text{CRE}(x^0\ |\ Y^*, P_\mathcal{X})=\underset{R^i\in \mathcal{R}^{\text{max}}_{Y^*}}{\text{argmin}}\ \ \text{cost}(x^0, R^i\ |\ P_\mathcal{X}).
    \vspace{-0.1cm}
\end{equation}
As shorthand, let $R^*=\text{CRE}(x^0\ |\ Y^*, P_\mathcal{X})$.
Depending on the choice of cost function, there may be many other inputs in $\mathcal{X}$ for which $R^*$ is the optimal (lowest-cost) counterfactual rule among the maximal-valid set $\mathcal{R}^{\text{max}}_{Y^*}$.
That is, there exists a \textit{group} of inputs which receive the \textit{same} CE as $x^0$.
We propose to describe this group of commonly-explained inputs using \textit{metarules}, drawn from the same class of interpretable rules $\mathcal{R}$. Retaining the terminology used above, we say that a metarule $M^i\in\mathcal{R}$ is valid if it only contains inputs for which $R^*$ is the optimal rule given $Y^*$ and $P_\mathcal{X}$,
\vspace{-0.1cm}
\begin{equation} \label{eq:metarule_is_valid}   
    \text{val}_{\text{meta}}(M^i) = \mathbbm{1}[\forall x {\in}  M^i, \text{CRE}(x\ |\ Y^*, P_\mathcal{X}){=}R^*],
    \vspace{-0.1cm}
\end{equation}
and maximal if it is not a subset of another valid metarule,
\begingroup
    \setlength\abovedisplayskip{0.1cm}
    \setlength\belowdisplayskip{0.0cm}
    \begin{multline}    
        \mathcal{M}^{\text{max}}_{R^*} = \{M^i \in \mathcal{R}\ |\ \text{val}_{\text{meta}}(M^i)\ \land \\
        (\nexists M^j \in \mathcal{R} : M^i \subset M^j \land \text{val}_{\text{meta}}(M^j))\}.
    \end{multline}
\endgroup

The addition of metarules enhances the explanatory power of counterfactual rules in at least two ways.
When presenting $R^*$ as the local explanation of $x^0$ to a user, it could be beneficial to accompany it with a member of $\mathcal{M}^{\text{max}}_{R^*}$ containing $x^0$ itself, as this provides background context on \textit{where else} $R^*$ is the optimal rule, and thus the robustness of the explanation to input perturbations \cite{mishra2021survey}.
Furthermore, the full set of counterfactual rules and metarules for an entire dataset provides a summary of the model's global recourse properties, which may assistive for model development, debugging and auditing \cite{ley2023globe}.
We explore this opportunity for global analysis in Section~\ref{sec:interpretability}.

\begin{figure*}
    \centering
    \includegraphics[width=\textwidth]{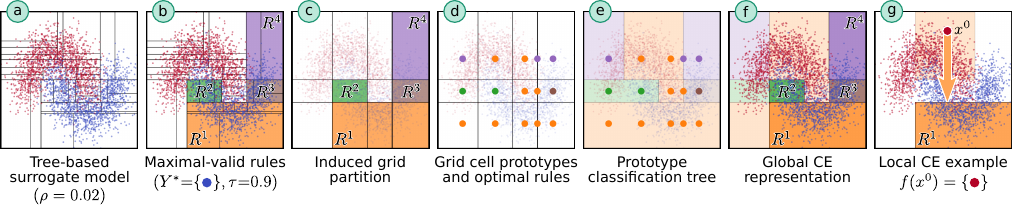}
    \vspace{-1.05cm}
    \caption{
    The seven steps of the \ours{} algorithm.
    }
    \label{fig:method}
    \vspace{-0.2cm}
\end{figure*}

\vspace{-0.1cm}
The rule/metarule formulation is extremely general and could be applied to any kind of input space, including images and text. Henceforth, however, we assume real vector input spaces $\mathcal{X}=\mathbb{R}^D$, spanned by features corresponding to intrinsically interpretable quantities, and consider rules and metarules with axis-aligned hyperrectangular geometry.
That is, each $R^i\in\mathcal{R}$ is defined by (potentially infinite) lower and upper bounds $l^i_d,u^i_d\in\mathbb{R}\cup\{-\infty,\infty\}$, $u^i_d\geq l^i_d$, along each feature $d\in\{1,\dots,D\}$.
Rules of this form can be seen as preserving feature-level interpretability as they can be expressed as a conjunction of single-feature terms, e.g. ``age $> 25$ and $\$30k<\text{income}\leq\$50k$''.
We note that one-hot encoded categorical features can be handled with a few additional constraints, discussed in Appendix \ref{app:categoricals_constraints}.

\vspace{-0.1cm}
The cost function in Equation \ref{eq:CRE} can also be defined in many reasonable ways depending on context.
Here, we consider
\vspace{-0.1cm}
\begin{equation} \label{eq:cost_fn}
    \text{cost}(x^0, R^i\ |\ P_\mathcal{X}) = \text{changes}(x^0, R^i) - \text{feasibility}(R^i),
    \vspace{-0.3cm}
\end{equation}
where
\vspace{-0.125cm}
\begin{equation} \label{eq:numchanges}
    \text{changes}(x^0, R^i) = \textstyle\sum_{d=1}^D\mathbbm{1}[(x^0_d\leq l^i_d) \lor (u^i_d < x^0_d)].
    \vspace{-0.1cm}
\end{equation}
This cost function prefers \textit{sparse} rules that require a small number of feature changes, thereby making them simpler to describe and use for recourse.
Sparsity objectives have significant precedence in the CE literature~\cite{guidotti2022counterfactual}.
It also prefers rules with higher feasibility under $P_\mathcal{X}$, which also follows prior work in hypothesising that counterfactuals in highly-populated regions can be more realistically used for recourse~\cite{poyiadzi2020face}.
Importantly, since $\text{changes}(\cdot, \cdot)$ is an integer and $\text{feasibility}(\cdot)\leq 1$, the first term always takes priority in cost calculations.

\vspace{-0.1cm}
Returning to the example in Figure~\ref{fig:page1}, the green and orange rules are maximal-valid for $Y^*=\{\bdot\}$ and any $\tau\leq 0.9$ and $\rho\leq 0.2$ (i.e. $10/50$ points). The inputs for which the orange rule is an optimal CE under Equation~\ref{eq:cost_fn} are contained in a single maximal-valid metarule ($x_1\leq 3$ and $x_2\leq 5$) while the green rule has two such metarules ($x_1\leq 3$ and $x_2> 5$, or $x_1>3$). An important property of textual representations of counterfactual rules (shown on the right)
 is that they differentiate between features that need to change (\textit{``change to...''}) and those that need to remain within a specified range (\textit{``while keeping...''}), both of which are required for the CE to inform reliable recourse for an end user. Notice how this leads the same rule to be expressed differently in different cases: satisfying the green rule requires changing feature $1$ in its first metarule, and feature $2$ in its second metarule.


\vspace{-0.125cm}
\section{
\ours{} \small{( \underline{T}rees for \underline{C}ounterfactual \underline{R}ule \underline{Ex}planation )}
} \label{sec:method}

We now describe \ours{}, an efficient, model-agnostic algorithm for generating valid (and approximately maximal) hyperrectangular rules and metarules for the cost function in Equation \ref{eq:cost_fn}.
It consists of seven steps, which are visualised for a toy example in Figure~\ref{fig:method}.
For simplicity, we assume here that all features are numerical,
and present refinements for handling one-hot encoded categoricals in Appendix~\ref{app:categoricals_algo}.

\vspace{-0.05cm}
\annot{a} Given a dataset of realistically-distributed inputs and associated model outputs, $\mathcal{D}=\{(x\sim P_\mathcal{X}, y=f(x))\}_{n=1}^N$, we first grow a tree-based \textit{surrogate} model, which can be either a random forest or a single decision tree.\footnote{We use a classification forest/tree if $f$ is a classifier, and a regression forest/tree if $f$ is a regressor.}
Each of the $L$ leaves and $L-1$ internal nodes of each tree equates to a hyperrectangle in $\mathcal{X}=\mathbb{R}^D$, whose bounds are determined by splits made at that node's ancestors. The tree growth algorithm optimises for a measure of \textit{purity} in the model outputs of data at every node, which closely aligns with our aim of finding accurate rules. We also use a stopping criterion to ensure that each leaf contains a minimum fraction $\rho \in (0,1]$ of the data in $\mathcal{D}$, thereby enforcing the feasibility constraint in Equation~\ref{eq:rule_is_valid}.
We then take all hyperrectangles corresponding to the nodes of the surrogate tree(s) as a candidate rule set $\mathcal{R}^\text{surr}$,
from which counterfactuals will be constructed.
In Figure~\ref{fig:method}, the surrogate is single tree which (for $\rho=0.02$) has $L=37$ leaves, yielding $|\mathcal{R}^\text{surr}| = 2L-1=73$ rules.


\vspace{-0.05cm}
\annot{b} From this step onwards, we assume that a set of target outputs $Y^*\subset Y$ and an accuracy threshold $\tau\in(0,1]$ have been specified.
We use these parameters (together with the guarantee that all rules in $\mathcal{R}^\text{surr}$ satisfy the feasibility constraint) to identify the maximal-valid rules,\footnote{
For some $(\rho,\tau)$ pairs, it is possible that $\mathcal{R}^{\text{max}}_{Y^*}=\emptyset$. In such cases, the algorithm should be run with different hyperparameters.
}
\begingroup
    \setlength\abovedisplayskip{0.05cm}
    \setlength\belowdisplayskip{0.15cm}
    \begin{multline}
        \mathcal{R}^{\text{max}}_{Y^*} = \{R^i \in \mathcal{R}^\text{surr}\ |\ \text{accuracy}(R^i) \geq \tau\ \land \\
        (\nexists R^j\in\mathcal{R}^\text{surr}:R^i \subset R^j \land \text{accuracy}(R^j) \geq \tau)\},
    \end{multline}
\endgroup
where the subset relation for hyperrectangles is defined as
\begin{equation} \label{eq:contains}
    R^i \subset R^j \equiv \textstyle\prod_{d=1}^D\mathbbm{1}[(l^j_d {\leq} l^i_d) {\land}  (u^i_d {\leq} u^j_d)] \land R^i\neq R^j
\end{equation}
and $\text{accuracy}(\cdot)$ is computed via Equation~\ref{eq:accuracy}, using $\mathcal{D}$ as a finite-sample approximation of $P_\mathcal{X}$. In the Figure~\ref{fig:method} example, we obtain $\mathcal{R}^{\text{max}}_{Y^*}=\{R^1,R^2,R^3,R^4\}$ for $Y^*{=}\{\bdot\}$, $\tau{=}0.9$.

\vspace{-0.05cm}
\annot{c} Next, we use all bounds occurring in the maximal-valid rules to partition the input space into a grid of hyperrectangular cells. That is, we find the set of unique bounds along each feature $d\in\{1,\dots,D\}$ (always including $\pm\infty$),
\vspace{-0.075cm}
\begin{equation}
    \text{bounds}(\mathcal{R}^{\text{max}}_{Y^*}, d)=\textstyle\bigcup_{R^i\in\mathcal{R}^{\text{max}}_{Y^*}}\{l^i_d,u^i_d\}\cup\{-\infty,\infty\},
    \vspace{-0.075cm}
\end{equation}
then construct cells using all possible combinations of consecutive pairs of bounds along all features. The worst-case size of the resultant grid is $(2|\mathcal{R}^{\text{max}}_{Y^*}|+1)^D$ cells, which in the Figure~\ref{fig:method} example is $27$, but the actual number is only $15$ here due to bound values being duplicated across rules.

\vspace{-0.1cm}
To understand why this grid partition is meaningful, consider the following reasoning:
\begin{itemize}
    \vspace{-0.4cm}
    \item For any two points in the same grid cell $C$, the set of features that must change to move from those points to each $R^i\in\mathcal{R}^{\text{max}}_{Y^*}$ must be the same, by virtue of the cell's construction from consecutive bounds in $\mathcal{R}^{\text{max}}_{Y^*}$.
    \vspace{-0.25cm}
    \item This means that the cost of each $R^i\in\mathcal{R}^{\text{max}}_{Y^*}$ under Equation~\ref{eq:cost_fn} must be constant throughout $C$.
    \vspace{-0.25cm}
    \item As a result, there exists an $R^{*C} \in \mathcal{R}^{\text{max}}_{Y^*}$ that has the lowest cost throughout $C$.
    \vspace{-0.25cm}
    \item Therefore, \underline{every grid cell is a valid metarule}.
\end{itemize}

\vspace{-0.35cm}
\annot{d} With the above in mind, the next step is to find the optimal rule for each grid cell $C$. An efficient way to do this is to pick an arbitrary point $x^C \in C$, which we call a \textit{prototype}, and use Equation~\ref{eq:CRE} (with Equation~\ref{eq:cost_fn} as the cost function) to find $R^{*C}=\text{CRE}(x^C\ |\ Y^*, P_\mathcal{X})$,\footnote{We break ties in the cost of two or more rules by taking the first in a fixed (but otherwise arbitrary) ordering of $\mathcal{R}$.} where $\mathcal{D}$ again serves as a finite-sample approximation of $P_\mathcal{X}$.
By definition, $R^{*C}$ will also be optimal throughout the rest of the cell. 
In Figure~\ref{fig:method}, we show
a prototype for each of the $15$ cells
and use colours to denote their optimal rules.

\vspace{-0.05cm}
\annot{e} At this point, we have identified the maximal-valid rules $\mathcal{R}^{\text{max}}_{Y^*}\subseteq\mathcal{R}^\text{surr}$ and valid metarules (i.e. grid cells) denoting their regions of optimality. However, the cells are unlikely to be maximal; Figure~\ref{fig:method} includes several instances of adjacent cells that could be merged to give a larger metarule with the same optimal rule. While one could develop a bottom-up algorithm for iteratively merging cells into larger metarules as the previous sentence implies, we instead pursue an efficient top-down approach. That is, we consider the set of cell prototypes, together with their optimal rule assignments, as a kind of labelled dataset, and grow \textit{another tree model} (specifically a CART classifier \cite{breiman2017classification}) to classify prototypes based on their labels. Crucially, we constrain this tree's growth algorithm to only consider split thresholds in $\text{bounds}(\mathcal{R}^{\text{max}}_{Y^*}, d)$ for each $d\in\{1,\dots,D\}$, and grow the tree to purity. The result is that every leaf of the tree is a union of cells sharing a common optimal rule, and hence is an (approximately maximal) valid metarule.
In the Figure~\ref{fig:method} example, the $15$ cells are aggregated into $7$ metarules: three for $R^1$, two for $R^4$, and one each for $R^2$ and $R^3$.

\vspace{-0.05cm}
\annot{f} Together, the rules and metarules globally characterise the counterfactual structure of $f$ for the given $(\rho, Y^*,\tau)$. The visual representation shown in Figure~\ref{fig:method} is only possible when $D=2$, but the textual form exemplified in Figure~\ref{fig:page1} is more scalable. As we show in Section~\ref{sec:interpretability}, the model is also amenable to the kind of regional and feature-level interpretability analysis that is normally possible with trees.

\vspace{-0.05cm}
\annot{g} In turn, the global rule/metarule structure trivially enables the explanation of a single instance $x^0:f(x^0)\not\in Y^*$ by identifying its containing metarule (in Figure~\ref{fig:method}, this leads to $R^1$ being returned as the CE). Since metarules are arranged in a conventional classification tree structure, highly optimised implementations can be leveraged to rapidly generate CEs for many instances in parallel.

\vspace{-0.1cm}
Steps \annot{b} -- \annot{e} can be repeated for any new $(Y^*, \tau)$ combination as required, to extract paired sets of rules and metarules from the same underlying surrogate.
Once this has been done once, the rule structures can be reused indefinitely for local explanation. This front-loading of computation makes \ours{} highly scalable to large datasets. In Appendix~\ref{app:complexity}, we provide a complexity analysis of each stage of the algorithm.
The step that is most computationally expensive in practice, \annot{d}, has complexity $O(\frac{DT^{D+1}}{\rho^{D+1}})$, which increases polynomially with higher numbers of trees in the surrogate model (denoted by $T$) and lower values of the hyperparameter $\rho$, and increases exponentially with the input space dimensionality $D$. Despite this exponential scaling, we find in Section~\ref{sec:expt_baselining} that \ours{} runtimes are orders-of-magnitude faster than baseline methods.

The lack of restriction on $Y^*$ (it can be any subset of $\mathcal{Y}$) makes \ours{} very flexible. If $f$ is a $K$-class classifier (i.e. $|\mathcal{Y}|=K$), we can handle both \textit{targeted} CE, in which $Y^*=\{y\}$ for some $y\in\mathcal{Y}$, and \textit{untargeted} CE, in which $Y^*=\mathcal{Y}\setminus\{y\}$.
For regression models, the CE problem can be formalised in an even greater diversity of ways \cite{spooner2021counterfactual}. We consider a simple treatment in Section~\ref{sec:regression}.


\begin{figure*}
    \centering
    \includegraphics[width=\textwidth]{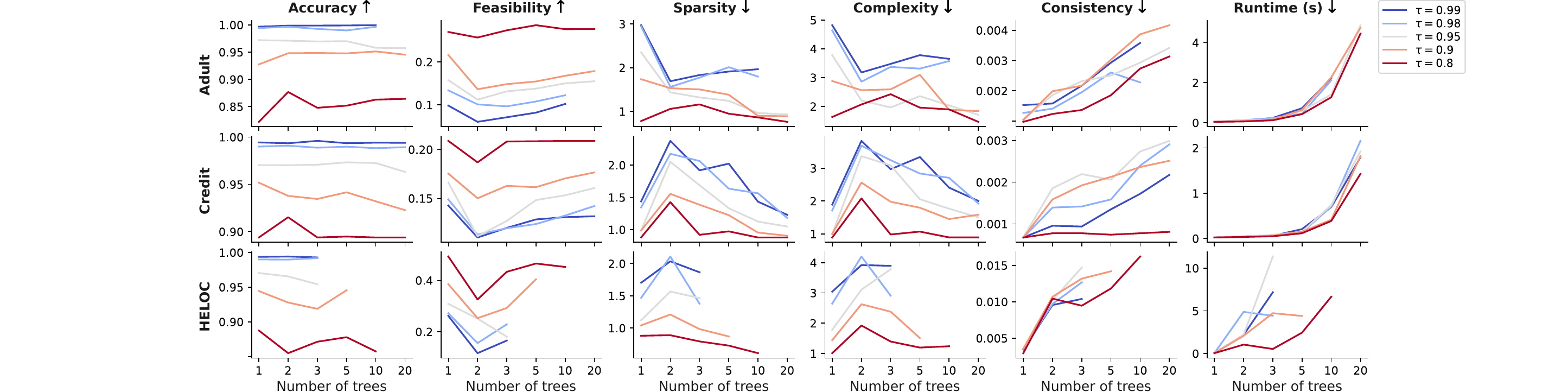}
    \vspace{-0.7cm}
    \caption{Performance of \ours{} as a function of the number of trees and $\tau$ (arrows indicate `better' direction for each desideratum).}
    \label{fig:hyperparam_subset}
\end{figure*}

\vspace{-0.15cm}
\section{Related Work}

As noted in the introduction, our work connects two strands of prior research. The first is concerned with finding group-level CE representations for sets of inputs, which may consist of single counterfactual points, \cite{warren2023explaining,carrizosa2024mathematical}, single vectors by which all inputs should be translated \cite{Kanamori2022counterfactual}, or single translation directions while allowing magnitudes to vary \cite{ley2023globe}.
Such aggregation provides a basis for high-level auditing of a model's counterfactual fairness properties and may enhance the trust and understanding of non-expert users \cite{warren2023explaining}.
The second strand aims to find diverse sets of CEs for single inputs, typically via gradient-based methods \cite{mothilal2020explaining,rodriguez2021beyond} 
or genetic algorithms \cite{dandl2020multi}. By providing fuller insight into a model's counterfactual options, it has been argued that diverse CEs improve actionability in the recourse setting \cite{wachter2017counterfactual}, although user studies suggest that presenting too many options can create \textit{``increased cognitive load that} [\textit{hinders}] \textit{understanding''} \cite{tompkins2022effect}. This drawback motivates providing compressed summaries of diverse CEs in the form of human-readable rules.


\vspace{-0.1cm}
Outside of the counterfactual context, rules have been used as factual explanations, denoting the sufficient conditions for a model to produce a given output \cite{ribeiro2018anchors,amoukou2022consistent}. To our knowledge, \lore{} \cite{guidotti2019factual} is the first method to combine such factual rules with counterfactual ones, describing ranges of feature value changes that (starting from a given reference input) would change the model output with high probability.
\citet{fernandez2020random} propose \rfocse{}, which finds similar counterfactual rules for the specific class of random forest classifiers, and demonstrate greatly improved computational speed over \lore{} (although the resultant rules have much lower feasibility).
Most related to our proposal is \ares{} \cite{rawal2020beyond}, which learns counterfactual rules for sets of inputs described by \textit{``subgroup descriptors''} and \textit{``inner rules''} (collectively analogous to our metarules). Rules and subgroups are jointly optimised to yield CEs that are accurate, incur low feature-changing costs, and cover as many instances as possible.
One drawback of \ares{}, alongside its extremely long runtimes \cite{ley2023globe}, is its use of arbitrary binning for numerical features. 
By extracting split thresholds from a tree-based surrogate model, our \ours{} method learns more adaptive rules while avoiding the need for binning.
In addition, \ares{}, \lore{} and \rfocse{} alike are only designed to work with binary classification models.
\ours{} natively handles both multi-class classifiers and regressors.

\vspace{-0.1cm}
Several works cited above leverage tree models, including \citet{Kanamori2022counterfactual} and both \lore{} and \rfocse{}. Our use is rather different, and in particular, the growth of a secondary tree for metarule aggregation is entirely novel.





\vspace{-0.15cm}
\section{Quantitive Experiments}

To evaluate \ours{}, we use it to generate counterfactual rules for unseen test data $\mathcal{D}_\text{test}{=}\{(x\sim P_\mathcal{X}, y{=}f(x))\}_{m=1}^M$ and score the results according to six desiderata.
The first of these are \textbf{feasibility} and \textbf{accuracy} (Equations~\ref{eq:feasibility} and~\ref{eq:accuracy}), both approximated by using $\mathcal{D}_\text{test}$ itself.
In rare cases that feasibility is $0$, we fall back to a default accuracy based on the marginal probability of $Y^*$ across $\mathcal{D}_\text{test}$.
We also report the \textbf{sparsity} of each rule (Equation~\ref{eq:numchanges}), as well as its \textbf{complexity}, which we define as the number of finite terms in its hyperrectangle bounds. We propose this desideratum because it equates to the number of expressions needed to describe a rule in text (fewer is better).
Our fifth desideratum is \textbf{consistency}, which is the number of unique rules returned across all inputs in $\mathcal{D}_\text{test}$ (divided by $|\mathcal{D}_\text{test}|$).
We suggest that having few unique rules is preferable because this implies robustness of the explanations to input perturbations. It also enables a compact representation of all counterfactuals for $\mathcal{D}_\text{test}$, which is a key motivation for group-level CE methods.
Finally, we report the \textbf{runtime} (on an \texttt{r6i.large} AWS instance) of all algorithmic variants and baselines. This is a crucial consideration for the deployment of explanation methods in practice.
Throughout this section, $f$ is an XGBoost model \cite{chen2016xgboost}, trained on $\mathcal{D}$ with \texttt{n\_estimators=50} and \texttt{max\_leaves=8}. However, \ours{} is model-agnostic, and we report similar results for a neural network model in Appendix~\ref{sec:app_mlp}.

\begin{figure*}
    \centering
    \includegraphics[width=\textwidth]{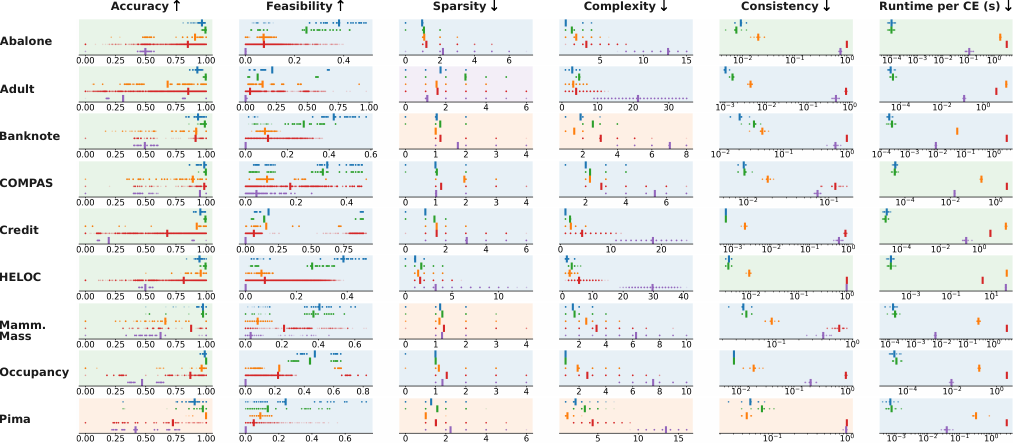}
    \vspace{-0.7cm}
    \caption{Comparative evaluation of \ourstaulo{}, \ourstauhi{}, \ares{}, \lore{} and \rfocse{} on nine binary classification datasets.}
    \label{fig:main_results}
\end{figure*}

\vspace{-0.2cm}
\subsection{Hyperparameter Study} \label{sec:hyperparam}

We begin by characterising the performance of \ours{} as a function of key hyperparameters, specifically the number of trees in the surrogate model ($\in\{1, 2, 3, 5, 10, 20\}$) and the accuracy threshold $\tau$ ($\in\{0.8, 0.9, 0.95, 0.98, 0.99\}$) while holding the feasibility threshold constant at $\rho=0.02$. We run this experiment on nine binary classification datasets (details in Appendix~\ref{app:datasets}), using $10$-fold cross-validation (CV) to split the datasets into train and test components $(\mathcal{D},\mathcal{D}_\text{test})$, and aggregate results across all folds.
We focus on the three largest datasets in Figure~\ref{fig:hyperparam_subset},
and report results for all nine datasets (and alternative $\rho$ values) in Appendix~\ref{app:hyperparam_full}.


\vspace{-0.1cm}
The first high-level trend to note is a clear and intuitive trade-off between the accuracy of returned rules on the one hand, and their feasibility, sparsity and complexity on the other. This trade-off is mediated by the accuracy threshold $\tau$. When higher thresholds are specified, this necessitates more specific rules with narrower bounds, which in turn sacrifices performance on the other three desiderata.

\vspace{-0.1cm}
The trends with tree count align less with our a priori expectations. One might assume that most desiderata would robustly improve as more trees are used, as this creates a larger pool of candidate rules to optimise over. For feasibility, sparsity and complexity, this usually appears to be true for $2$-$20$ trees,\footnote{Some results for higher tree counts are absent here due to a manually-imposed cell limit being reached; see Appendix~\ref{app:hyperparam_full}.} but in many cases (most visibly for Credit) using a single tree markedly outperforms using two. We believe the outsized performance of single trees is due to their use of the entire dataset $\mathcal{D}$ for growth rather than bootstrap samples, as in typical random forest implementations.

\vspace{-0.1cm}
There is no consistent trend towards more accurate rules as tree count increases, and consistency reliably worsens, as having more valid rules to select from leads to more fragmented results. The runtime required to learn all rules and metarules also increases superlinearly. Together with the ambiguous and inconsistent trends in the other desiderata, this provides a strong practical reason for initially running the \ours{} algorithm with a single surrogate tree, and only deviating from this if unsatisfactory results are obtained.
For this reason, we use a single tree in all remaining experiments. Specifically, we consider two variants with $\tau=0.9$ (\ourstaulo{}) and $\tau=0.99$ (\ourstauhi{}).

\vspace{-0.2cm}
\subsection{Baseline Comparison} \label{sec:expt_baselining}

We now present our main results, which compare \ourstaulo{} and \ourstauhi{} to the three relevant baselines from prior work: \ares{}, \lore{} and \rfocse{} (see Appendix~\ref{app:baselines} for baseline details). We use the same nine datasets and 
CV setup as above, and evaluate the rules returned by each method using our six key desiderata. Figure~\ref{fig:main_results} shows the distribution of results for each desideratum, dataset and method. Thick vertical lines denote mean values, and subplot background colours indicate the best-performing method.

\vspace{-0.1cm}
Moving left-to-right through the columns, we immediately encounter a positive result. \ourstauhi{} (the variant with the more stringent accuracy threshold $\tau$) returns the most accurate rules on eight out of nine datasets, only being narrowly beaten by \ares{} on Pima. As expected, \ourstaulo{} returns somewhat less accurate rules, but this is counterbalanced by their superior feasibility: this variant of our method returns the most feasible rules on \textit{every} dataset. \ourstauhi{} still performs well on feasibility, ranking second on seven datasets, often with a sizeable gap to the third-ranked baseline.
This indicates that our method achieves strong compromises on the accuracy-feasibility trade-off.
\rfocse{}'s poor feasibility results are notable; it frequently returns rules so specific that they contain zero instances from the test set.

\vspace{-0.1cm}
The methods perform more similarly in terms of sparsity.
In most cases, they return rules requiring just one or two features to be changed.
Overall, we feel comfortable in declaring \ourstaulo{} a narrow winner on this desideratum, as it ranks highest on five of nine datasets. This is true to a greater extent for complexity, where \ourstaulo{} ranks best on seven datasets, and never outside the top two. The only consistently strong baseline on both sparsity and complexity is \ares{}. \ourstauhi{} is fairly competitive, outperforming at least two baselines in most cases, but is always beaten by \ourstaulo{}. This reinforces our earlier observation that more accurate rules come at the cost of other desiderata.

\vspace{-0.1cm}
On consistency, which measures the fraction of unique rules as an indicator of robustness and explanatory simplicity, \ours{} outperforms all baselines for eight of nine datasets. Consistencies of $10^{-1}$ to $10^{-3}$ indicate the same rule is returned for tens or hundreds of test instances. Once again, \ares{} is the only close competitor, which is intuitive as it also learns aggregated group-level explanations.
\lore{} and \rfocse{} perform no such aggregation, so any repetition of rules (consistency $<10^0=1$) is coincidental.

\vspace{-0.1cm}
The disparity of runtimes is the most stark, varying by $3$-$4$ orders of magnitude between both \ours{} variants (which are always fastest) and the slowest baseline. To illustrate the strength of this result: the total runtime for \textit{all} test inputs in \textit{all} CV folds of \textit{all} datasets is
$4.16s$ and $4.52s$ 
for \ourstaulo{} and \ourstauhi{} respectively, which is $13\times$ less than the mean runtime \textit{per input} of \ares{} on HELOC.
\ours{} is at least an order of magnitude faster than \rfocse{} on every dataset, which is notable because this baseline's speed is hailed as a key advantage by its authors.

\vspace{-0.2cm}
\subsection{Counterfactual Distance Evaluation}

An apparent drawback of \ours{} is that it does not optimise for rules that require small magnitudes of feature changes to the original input, as measured by metrics such as the Manhattan or Euclidean distance to the closest point satisfying the rule. This desideratum, which is distinct from sparsity, is very common in the literature
\cite{karimi2022survey},
and is included to some extent in all three baselines. We can justify our omission: since most distance metrics vary continuously, their inclusion would prevent us from finding non-infinitesimal hyperrectangular metarules for which a single counterfactual rule is guaranteed to be optimal. However, given its ubiquity, it is important to know how our method compares to baselines on this desideratum.

\begin{figure}[H]
    \centering
    \includegraphics[width=\columnwidth]{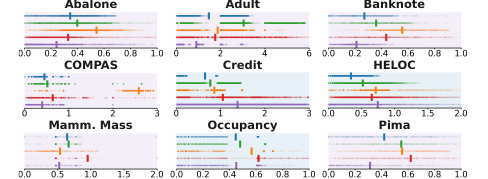}
    \vspace{-0.75cm}
    \caption{Distribution of counterfactual distances for all methods.}
    \label{fig:point_distances}
\vspace{-0.2cm}
\end{figure}

\vspace{-0.1cm}
The results, shown in Figure~\ref{fig:point_distances}, are encouraging. For all nine datasets,
we report the distribution of distances between each test input and the closest point in the rule returned by each method, where distance is measured as the \textit{total percentile shift}~\cite{pawelczyk2020learning}.
Despite \ours{} making no explicit attempt to minimise counterfactual distances, it performs comparably to the baselines, with both \ourstaulo{} and \ourstauhi{} outperforming \ares{} and \lore{} on seven of the nine datasets. In most cases, \rfocse{} is strongest on this metric, but \ourstaulo{} actually does best on three datasets. The fact that \ourstaulo{} always yields lower distances than \ourstauhi{} suggests a reason for these positive results: by optimising for feasibility (which \ourstaulo{} does to a greater extent than \ourstauhi{}), we indirectly incentivise rules that occupy as much volume in the input space as possible, which in turn reduces the expected distance to any other point in the space.

\vspace{-0.15cm}

\vspace{-0.35cm}
\subsection{Regression Example} \label{sec:regression}

All three baselines are designed exclusively for binary classification, but \ours{} can operate in a much wider range of contexts, including regression. To demonstrate this, we evaluate the method on an XGBoost regression model for the Wine Quality dataset \cite{misc_wine_quality_186}. As discussed above, the CE problem for regression has many valid formulations, but we use a simple approach that partitions the output space into `low' and `high' halves using the mean model output $\mu$ across $\mathcal{D}_\text{test}$. Concretely, we set $Y^*=(\mu, \infty)$ if $f(x^0) \leq \mu$ and $Y^*=(-\infty,\mu]$ otherwise. As in binary classification, this requires steps \annot{b} -- \annot{e} of the \ours{} algorithm to be completed twice: once for low-to-high counterfactuals, and once for high-to-low.

\begin{figure}[H]
    \centering
    \includegraphics[width=\columnwidth]{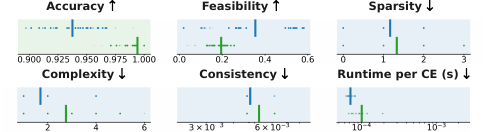}
    \vspace{-0.75cm}
    \caption{Evaluating \ourstaulo{} and \ourstauhi{} for regression.}
    \label{fig:wine_quality}
\vspace{-0.2cm}
\end{figure}

\vspace{-0.15cm}
The performance of \ourstaulo{} and \ourstauhi{} on our six key desiderata are shown in Figure~\ref{fig:wine_quality}. The high-level outcomes are consistent with the binary classification setting, insofar as \ourstauhi{} reliably returns more accurate rules, at the expense of somewhat worse feasibility, sparsity and complexity. Both variants produce CEs in $\approx0.1ms$ per instance, similar to comparably-sized datasets in Figure~\ref{fig:main_results} (COMPAS, HELOC). This experiment thus provides good evidence that \ours{} generalises well to regression.
\vspace{-0.15cm}

\setcounter{figure}{7}
\begin{figure*}[!b]
    \centering
    \includegraphics[width=\textwidth]{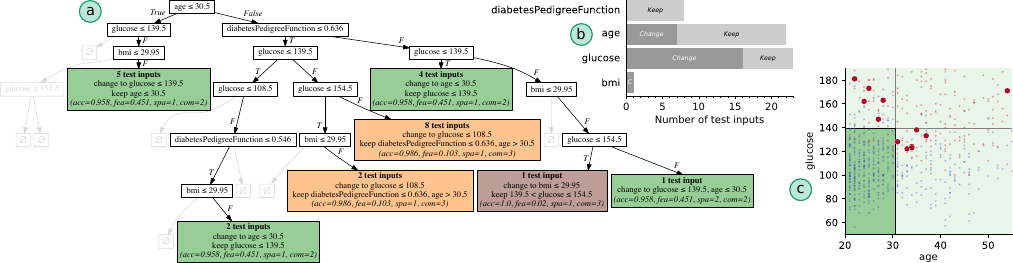}
    \vspace{-0.7cm}
    \caption{Rules and metarules for the Pima diabetes dataset, with analysis for a particular sample of 23 test inputs.}
    \label{fig:subgroup_and_features}
\end{figure*}

\vspace{-0.35cm}
\section{Qualitative Analysis} \label{sec:interpretability}

Having provided evidence of the promising quantitative performance of \ours{, we now briefly explore the qualitative structure of the rule-based counterfactuals it provides. 

\vspace{-0.2cm}
\subsection{Global CE Summary}

We begin by examining the rules and metarules learnt by \ourstaulo{} for $Y^*=\{high\ income\}$ on the Adult dataset (CV fold 7). Figure~\ref{fig:global_ce} depicts this information as a tree diagram, as an alternative to the purely textual form exemplified in Figure~\ref{fig:page1}.
In this case, our method learns a total of five metarules (leaves of this tree) for two distinct counterfactual rules (green/orange colouring), whose accuracy, feasibility and complexity are shown in italics. We also show the sparsity, which can differ between metarules for the same rule.\footnote{Strictly speaking, this is the \textit{worst-case} sparsity for each (rule, metarule) pair. It is possible to satisfy a metarule and need to change fewer than the stated number of features (for example, if a person is already married).} This diagram provides a complete global summary of the counterfactual rules learnt for this dataset, which describe options for changing the model output from $low\ income$ to $high\ income$ in terms of a range of alternative values for an individual's age, education level, marital status and working hours per week. We can also see that the rule returned for any given individual depends on their current age and working hours.

\setcounter{figure}{6}
\vspace{-0.15cm}
\begin{figure}[H]
    \centering
    \includegraphics[width=\columnwidth]{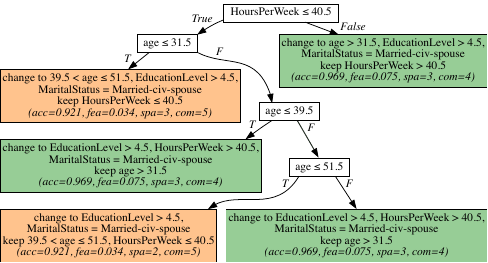}
    \vspace{-0.75cm}
    \caption{Rules and metarules for the Adult dataset.
    }
    \label{fig:global_ce}
\vspace{-0.2cm}
\end{figure}

\vspace{-0.35cm}
\subsection{Local CE Example and Metacounterfactual}


Remaining with the above example, consider an unmarried individual aged $45$, who has no degree and works $35$ hours per week, and for whom the model output is $low\ income$. They may wish to know why the output is not $high\ income$ instead. Running their features through the metarule tree leads to the bottom-left of Figure~\ref{fig:global_ce}, and hence to the yellow counterfactual rule. That is, the model would output $high\ income$ with high probability if this individual were married and had an education level of $5$ (Bachelor's degree) or higher, provided their age remained in the $40$-$51$ range and their working hours remained at $40$ or fewer.

\vspace{-0.1cm}
Now consider what would happen if this individual were $55$ years old instead of $45$.
They would be led to the bottom-right of Figure~\ref{fig:global_ce}, and hence to the green metarule. Now, in order to be predicted as $high\ income$ with high probability, they would need to make the same feature changes as before (married, education level of $5$ or higher) \textit{and} work $41$ hours per week or more. The reason for this change, which likely results from trends in the training data reflected in the model, can be pinpointed to their age now being $52$ or higher. Such \textit{metacounterfactual} reasoning (i.e. concerning the input changes required to yield an alternative CE) offers an interesting perspective on the issue of explanation robustness \cite{mishra2021survey}, and provides a basis for understanding changes in recourse recommended for individuals as a result of natural changes in features such as their age.
We believe this basic idea warrants deeper investigation in future.

\vspace{-0.2cm}
\subsection{Dataset-level Analysis}

Figure~\ref{fig:subgroup_and_features} \annot{a} shows another set of rules and metarules learnt by \ourstaulo{}, in this case for $Y^*=\{no\ diabetes\}$ on the Pima diabetes dataset (CV fold 8). Here, we indicate where 23 unseen test inputs (for which the model output is $has\ diabetes$) fall in the metarule tree. We take the opportunity to collapse any parts of the tree where no data reside, which allows us to focus on the information that is relevant to this particular dataset. The test inputs are distributed between seven metarules for three distinct counterfactual rules. The most common metarule denotes individuals aged 31 or older, with a diabetes pedigree function of $0.636$ or below and a plasma glucose concentration exceeding $154.5$. For these individuals, it is predicted that the model would output $no\ diabetes$ with high probability if their glucose levels were reduced to $108.5$ or below (and their age and diabetes pedigree function value remained in the same range).

\vspace{-0.1cm}
High-level counterfactual summary statistics can be derived for this test set, such as Figure~\ref{fig:subgroup_and_features} \annot{b}, which shows the number of inputs for which each feature is included in the returned rule (\textit{``change''} and \textit{``keep''} conditions are plotted separately). This highlights the importance of glucose concentration for the diabetes diagnosis problem: the glucose feature must either be changed or kept within a specified range for all 23 test inputs (change: 16, keep: 7).

\vspace{-0.11cm}
12 of the test inputs fall in metarules for which the green counterfactual rule (glucose $\leq 139.5$ and age $\leq 30.5$) is optimal. This rule only includes two features, so we can represent it via a 2D plot similar to those in Figures~\ref{fig:page1} and~\ref{fig:method}. This is shown in Figure~\ref{fig:subgroup_and_features} \annot{c}, which also includes the values of these features for the training data $\mathcal{D}$ (small points) and the 12 test inputs themselves (large points). The five individuals aged 30 or below require different minimum glucose reductions to satisfy the rule, ranging from $7.5$ in the lowest case to $41.5$ in the highest. Excluding one outlier, those older than 30 already have glucose levels below the required threshold, and thus would be likely to receive a $no\ diabetes$ prediction if only they were a few years younger. A plot of this type may help to illustrate the differing implications of counterfactual rules for different individuals.


\vspace{-0.25cm}
\section{Conclusion}

We have introduced a fast, model-agnostic method for learning accurate and feasible counterfactual rules, alongside metarules for choosing between them, thereby enabling both local (individual) and global (group-level) CE. We have demonstrated the method's efficacy on benchmark classification and regression datasets with a mix of numerical and categorical features, and shown strong performance relative to baselines on a range of CE desiderata.

\vspace{-0.11cm}
A limitation of our current implementation is that is tied to a particular cost function and lacks the facility for domain-specific actionability constraints. This can lead it to return non-actionable counterfactuals (e.g. requiring an individual to lower their age).
Although such constraints may not be relevant for understanding a model's bias and fairness, they are crucial for recourse~\cite{karimi2022survey}. We believe that the method could be generalised to handle actionability and alternative costs without major changes. Also important is the question of tree count and other hyperparameters for surrogate tree growth. While we found single-tree surrogates to be notably effective in our experiments, this issue warrants deeper theoretical and empirical investigation.





\vspace{-0.25cm}
\section*{Impact Statement}

\begin{small}
This paper presents a method for explaining the outputs of black box AI models and summarising recourse options for both individuals and groups using human-interpretable rules. As with many proposals made within the wider XAI field, the responsible deployment of more mature versions of such a technology could have a positive societal impact, enabling more understandable, trustworthy and user-friendly AI systems in deployment.
\par
\end{small}

\vspace{-0.25cm}
\section*{Disclaimer}

\begin{small}
This paper was prepared for informational purposes by the Artificial Intelligence Research group of JPMorgan Chase \& Co and its affiliates (“J.P. Morgan”) and is not a product of the Research Department of J.P. Morgan.
J.P. Morgan makes no representation and warranty whatsoever and disclaims all liability, for the completeness, accuracy or reliability of the information contained herein.
This document is not intended as investment research or investment advice, or a recommendation, offer or solicitation for the purchase or sale of any security, financial instrument, financial product or service, or to be used in any way for evaluating the merits of participating in any transaction, and shall not constitute a solicitation under any jurisdiction or to any person, if such solicitation under such jurisdiction or to such person would be unlawful.
\par
\end{small}







\bibliography{bibliography}
\bibliographystyle{icml2024/icml2024}

\newpage
\appendix
\onecolumn

\section{Handling of One-hot Encoded Categorical Features}

\subsection{Well-formedness Constraints on Rules} \label{app:categoricals_constraints}

In one-hot encoding, a categorical feature $c$ taking $D_c$ possible values is mapped to a length-$D_c$ vector in which one `hot' element is equal to $1$ and the remaining `cold' elements are equal to $0$. We can therefore represent an arbitrary mix of numerical and categorical features as a single real vector space $\mathbb{R}^D$, of which a subspace $S_c\subseteq\{1,\dots,D\}:|S_c|=D_c$ represents each categorical $c$ and thus only contains $0$ and $1$ values.

As discussed in the main paper, a hyperrectangular rule $R^i$ is defined by lower and upper bounds $l^i_d,u^i_d\in\mathbb{R}\cup\{-\infty,\infty\}$ along each feature $d\in\{1,\dots,D\}$. For numerical features, these bounds are unconstrained (except that $u^i_d\geq l^i_d$), but for each categorical feature $c$, the following conditions must hold for the rule to be well-formed:
\begin{itemize}
    \vspace{-0.1cm}
    \item $R^i$ must either specify a single hot category $\text{hot}^i_c\in S_c$ or up to $D_c-1$ cold categories $\text{cold}^i_c\subset S_c$,\footnote{$\text{cold}^i_c$ can be the empty set, in which case the rule effectively ignores feature $c$.} but never multiple hot categories (which would be impossible to satisfy) and never a mixture of hot and cold (which would be over-specified; a single hot category fully determines the feature's value).
     \vspace{-0.1cm}
    \item In the one-hot case, $l^i_{\text{hot}^i_c}$ must be set to a value in $[0, 1)$. In practice, we use $0.5$.
     \vspace{-0.1cm}
    \item In the multi-cold case, for each $d\in \text{cold}^i_c$, $u^i_d$ must be set to a value in $[0, 1)$. As above, we use $0.5$.
     \vspace{-0.1cm}
    \item All other lower and upper bounds for each $d\in S_c$ must be set to $-\infty$ and $\infty$ respectively.
     \vspace{-0.1cm}
\end{itemize}

As a result, for any given point $x^0\in \mathbb{R}^D$, each categorical feature $c$ can contribute a value of either $0$ or $1$ to the $\text{changes}(x^0, R^i)$ calculation in Equation~\ref{eq:numchanges}:
\begin{itemize}
    \vspace{-0.1cm}
    \item In the one-hot case where $x^0_{\text{hot}^i_c}=1$, the contribution is $0$ because $\mathbbm{1}[(1\leq 0.5) \lor (\infty < 1)]=0$.
     \vspace{-0.1cm}
    \item In the one-hot case where $x^0_{\text{hot}^i_c}=0$, the contribution is $1$ because $\mathbbm{1}[(0\leq 0.5) \lor (\infty < 0)]=1$.
     \vspace{-0.1cm}
    \item In the multi-cold case where $\nexists d\in \text{cold}^i_c: x^0_d=1$, the contribution is $0$ because $\mathbbm{1}[(0\leq -\infty) \lor (0.5 < 0)]=0$.
     \vspace{-0.1cm}
    \item In the multi-cold case where $\exists d\in \text{cold}^i_c: x^0_d=1$, the contribution is $1$ because $\mathbbm{1}[(1\leq -\infty) \lor (0.5 < 1)]=1$.
\end{itemize}

\subsection{Algorithmic Refinements} \label{app:categoricals_algo}

The \ours{} algorithm includes several minor adjustments for the correct handling of categorical features.

\subsubsection{Rule Simplification on Extraction from the Surrogate}

During the greedy growth process of a standard classification or regression tree, it is possible to create branches with cold split outcomes for one or more categories of a categorical feature, followed by (lower down in the branch) a hot split outcome for another category. As discussed above, the presence of a hot category makes the specification of cold categories redundant, so when extracting rules from the surrogate in step \annot{a}, we apply a post hoc correction to all extracted rules to set the corresponding bounds to $\infty$. Secondly, while a rule specifying $D_c-1$ cold categories for a categorical feature $c$ is technically well-formed, it is equivalent to the simpler and more direct specification of the remaining hot category. We replace any such `all-but-one-cold' rules with their equivalent one-hot representations.

\subsubsection{Modified Subset Relation}

A more subtle correction is needed for the hyperrectangle subset relation in Equation~\ref{eq:contains}, which is used to identify maximal-valid rules in step \annot{b}. From a semantic perspective, a rule $R^j$ that is multi-cold for some categorical $c$ is less specific than 
another rule $R^i$ that is one-hot with $\text{hot}^i_c\not\in\text{cold}^j_c$ (assuming equal bounds on all other features), since the latter is a special case of the former. $R^i$ should therefore be considered a subset of $R^j$, but the way the bounds of one-hot rules are specified above (i.e. $\pm\infty$ for all $d\in S_c\setminus\{\text{hot}^i_c\}$) leads this not to be the case. To fix this, we temporarily modify all one-hot rules to have explicit upper bounds of $0.5$ for all cold categories, which equates to applying the following function:
\begin{equation}
    \hat{u}^i_d=\left\{\begin{array}{ll}
        0.5 & \text{if}\ \ \exists\ c : (d\in S_c)\land(\exists\ d'\in S_c\setminus \{d\}:l_{d'}^i=0.5), \\
        u^i_d & \text{otherwise.}
    \end{array}\right.
\end{equation}
The strict subset relation $R^i \subset R^j$ is then defined exactly as in Equation~\ref{eq:contains}, except using $\hat{u}^i_d$ and $\hat{u}^j_d$ instead of $u^i_d$ and $u^j_d$.

\subsubsection{Skipping of Impossible Grid Cells}

In step \annot{c}, the bounds of the maximal-valid rules are used to partition the input space into a grid of hyperrectangular cells. In the presence of categoricals, a large proportion of these cells can never be occupied by any real input point, because they specify multiple (or zero) hot categories for the same feature. Consider the following minimal example of an input space consisting of a single binary feature, which is partitioned into a $2\times 2$ grid:
\begin{figure}[H]
    \centering
    \includegraphics[width=1.7in]{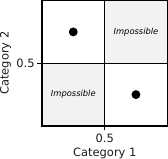}
    \label{fig:impossible_cells}
    \vspace{-0.3cm}
\end{figure}
The two cells on the off-diagonal can be occupied by real data as they specify one of the two categories being hot, but the other two cells are impossible as they specify either both or neither. Since we are only interested in finding CEs for inputs that are possible in practice, we skip all such impossible cells when generating prototypes. This delivers a significant efficiency saving: the maximum factor by which a categorical feature $c$ can increase the size of the grid is reduced from $2^{D_c}$ ($=4$ in the above example) to $D_c$ ($=2$).




\section{Computational Complexity Analysis} \label{app:complexity}

Let $D$ be the dimensionality of the input space, $N$ be the size of the dataset used to grow the surrogate model and $T$ be the number of trees in the surrogate model. The following is a computational complexity analysis of each step of the \ours{} algorithm (excluding step \annot{f}, which describes no specific computation):

\annot{a} This step consists of growing standard classification or regression trees with an early-stopping criterion enforcing that each leaf has a minimum fraction $\rho\in(0,1]$ of the data. This creates a maximum of $L_{\max}=\text{ceil}(1/\rho)$ leaves per tree via $L_{\max}-1$ splitting operations. Since each such operation involves searching over $O(DN)$ possible splits of the data, the complexity of tree growth is $O(DN(L_{\max}-1))=O(\frac{DN}{\rho})$. This implies a complexity of $O(\frac{TDN}{\rho})$ when growing $T$ independent trees to form a random forest, which will be somewhat reduced in typical implementations where each tree only uses a subset of features.

\annot{b} The greatest number of valid rules extracted from a single tree, denoted by $V$, will tend to increase as the hyperparameter $\tau\in(0,1]$ is decreased, but is upper-bounded by the maximum total number of rules: $V\leq 2L_{\max}-1=2\text{ceil}(1/\rho)-1$. Evaluating the pairwise subset relation to find the maximal-valid rules involves comparing each pair of rules on each feature, so for $T$ trees is $O(D(VT)^2)=O(D((2\text{ceil}(1/\rho)-1)T)^2)=O(\frac{DT^2}{\rho^2})$.

\annot{c} Let $M\leq V$ be the greatest number of maximal-value rules extracted from a single tree. Because these rules have been filtered by the subset relation, $M$ is upper-bounded by the maximum number of leaves: $M\leq L_{\max}=\text{ceil}(1/\rho)$. In the worst-case grid size scenario described in the paragraph for this step in Section~\ref{sec:method}, we must create $O((MT)^D)=O(\frac{T^D}{\rho^D})$ grid cell prototypes, which involves computing $O(\frac{DT^D}{\rho^D})$ individual feature values.

\annot{d} Using Equations \ref{eq:CRE} and \ref{eq:cost_fn} to identify the optimal rule is $O(DMT)=O(\frac{DT}{\rho})$ for each prototype, and hence is $O(\frac{T^D}{\rho^D}\times \frac{DT}{\rho})=O(\frac{DT^{D+1}}{\rho^{D+1}})$ for all prototypes.

\annot{e} This step involves growing another standard classification tree, but this time without an early-stopping criterion and with the $O(\frac{T^D}{\rho^D})$ prototypes as the points to be classified rather than the original dataset. The complexity of this process depends on how balanced the classification problem is, but has been generically estimated as $O(D[\text{num\ points}]\log [\text{num\ points}])$ per tree \cite{sani2018computational}.
This evaluates to $O(D\frac{T^D}{\rho^D}\log (\frac{T^D}{\rho^D}))=O(D^2\frac{T^D}{\rho^D}(\log T-\log\rho))$.

\annot{g} Finding the local explanation for a single instance involves propagating that instance through the metarule tree structure. In the worst case, this involves a number of feature comparisons equal to the depth of this tree, which is $O(\log ([\text{num\ points}]))=O(\log(\frac{T^D}{\rho^D}))=O(D(\log T-\log\rho))$ if the prototype classification problem in step \annot{e} is relatively balanced.

In practice, we find that \annot{d} is most computationally expensive step of the algorithm in all experimental settings studied in this paper. It is $O(\frac{DT^{D+1}}{\rho^{D+1}})$, which increases polynomially with higher tree counts $T$ and lower values of the hyperparameter $\rho$, and increases exponentially with the input dimensionality $D$.

Steps \annot{a}, \annot{e} and \annot{g} can all leverage highly-optimised decision tree implementations so have a small impact on the total runtime. Although we made an effort to implement the other steps efficiently using parallelised array operations, it is likely that further optimisation is possible.

\section{Dataset Details} \label{app:datasets}

Our experimental setup is somewhat inspired by that used in the \rfocse{} paper \cite{fernandez2020random}, and we retain eight out of the $10$ public-access datasets studied there. We remove Post-Operative Patient and Seismic Bumps due to their small size ($86$ instances) and extreme class imbalance ($0.07$) respectively. In their place, we add Home Equity Line of Credit (HELOC) and Wine Quality, the latter of which is a regression dataset to serve as a demonstration of our method's suitability for that context (see Section~\ref{sec:regression}). Otherwise, our data preprocessing pipeline is very similar to that of \citet{fernandez2020random}. Details of the $10$ datasets used in our experiments are as follows (\# Inst. = number of instances, \# Feat. = number of features, \# Cat. = number of categorical features, Class Balance = proportion of instances with the positive class):



\begin{table}[H]
\small
\centering
\begin{tabular}{lcccccc}
\textbf{Short Name} & \textbf{Full Name}             & \textbf{Citation}        & \textbf{\# Inst.}    & \textbf{\# Feat.} & \textbf{\# Cat.} & \textbf{Class Balance} \\
\hline
Abalone             & Abalone                        & \cite{misc_abalone_1}    & 4177                 & 8                 & 1                & 0.50$^{*}$                       \\
Adult               & Adult / Census Income          & \cite{misc_adult_2}      & 30,718               & 11                & 5                & 0.25                       \\
Banknote            & Banknote Authentication        & \cite{misc_banknote_authentication_267} & 1372  & 4                 & 0                & 0.44                       \\
COMPAS              & COMPAS Recidivism Racial Bias  & \cite{compas}            & 5278                 & 5                 & 3                & 0.53                       \\
Credit              & Default of Credit Card Clients & \cite{misc_default_of_credit_350} & 29,623      & 14                & 3                & 0.78                       \\
HELOC               & Home Equity Line of Credit     & \cite{heloc}             & 9871                 & 23                & 0$^{**}$          & 0.52                       \\
Mamm. Mass          & Mammographic Mass              & \cite{misc_mammographic_mass_161} & 830         & 5                 & 2                & 0.51                       \\
Occupancy           & Occupancy Detection            & \cite{misc_occupancy_detection__357} & 2665$^{***}$ & 5              & 0                & 0.64                       \\
Pima                & Pima Indians Diabetes          & \cite{smith1988using}    & 768                  & 8                 & 0                & 0.65                       \\
Wine Quality        & Wine Quality                   & \cite{misc_wine_quality_186} & 6497$^{****}$    & 12$^{****}$       & 1$^{****}$        & N/A
\end{tabular}
\end{table}

\vspace{-0.4cm}
$^{*}$ Abalone is natively a regression dataset, but \citet{fernandez2020random} transform it into a classification one by splitting the target variable at the median (hence the equal class balance).

$^{**}$ Strictly speaking, the \texttt{MaxDelq2PublicRecLast12M} and \texttt{MaxDelqEver} features are categoricals, mapping onto strings such as ``120+ Days Delinquent'' and ``Never Delinquent'' (see \url{https://docs.interpretable.ai/stable/examples/fico/}). However, the $0$-$9$ numerical encoding used in the dataset file imposes a semantically meaningful ordering on the categories, allowing these features to be treated as numerical in practice.

$^{***}$ This actually appears to be only the test split of the original dataset, but is the one used in \citet{fernandez2020random}'s GitHub repository at \url{https://github.com/rrunix/libfastcrf}, so we retain it here.

$^{****}$ We concatenate the red and white wine subsets of the original dataset and add a binary categorical feature to indicate the colour of each instance. The $11$ original features are all numerical.

\newpage
\section{Extended Hyperparameter Study} \label{app:hyperparam_full}

Figure~\ref{fig:hyperparam_full} (overleaf) shows full results for the surrogate tree count and accuracy threshold ($\tau$) hyperparameter study across all nine binary classification datasets.

Before turning to the desiderata, the leftmost columns consider two reasons why our \ours{} implementation may fail to return rules. Firstly, no valid rules may be found. In general, this becomes more likely for higher values of $\tau$, as it becomes harder to find rules that satisfy both the feasibility and accuracy criteria. Its likelihood decreases as more surrogate trees are used (more trees means more chances of finding a valid rule), with the notable exception of a single tree. As discussed in the main paper, we believe the strong performance of the single-tree case is due to its use of the entire dataset rather than bootstrap samples as in a random forest. We note that this failure mode never occurs for six out of nine datasets, indicating that it is not always an issue, even for very high $\tau$ values.

Alternatively, the algorithm might violate a self-imposed limit on the number of grid cells (we use $1e^5$), which we include to pre-empt and prevent long runtimes. Again, this failure mode does not occur for all datasets, but when it does, it becomes markedly more likely as the number of surrogate trees increases. This is because more trees tend to yield more maximal-valid rules, and thus more unique boundaries from which to construct the cell grid. It also tends to increase with $\tau$, as higher values of this accuracy threshold yield a larger number of more specific maximal-valid rules.

The remaining columns focus on cases for which the algorithm does not fail by either of the two modes described above. The key observations made in Section~\ref{sec:hyperparam} continue to apply across this larger suite of datasets, including:
\begin{itemize}
    \item The $\tau$-mediated trade-off between accuracy on the one hand, and feasibility, sparsity, complexity and (to a lesser extent) consistency is clear in the vast majority of cases.
    \item The trends with tree count are less straightforward. Feasibility, sparsity and complexity usually improve as the number of trees is increased from $2$-$20$, although in many cases using a single tree markedly outperforms using two. Consistency, on the other hand, worsens with tree count, as more unique maximal-valid rules exist.
    \item The trend of accuracy with tree count is either flat or, in some cases, actually decreasing. We believe that this is due to our use of accuracy as a minimum threshold for validity, rather than an objective to be optimised. Having more rules to choose from increases the optimisation space for the other desiderata of feasibility and sparsity, which (due to the aforementioned trade-off) indirectly leads to somewhat less accurate rules being returned.
    \item The runtime required to learn all rules and metarules increases superlinearly with tree count, which provides a strong motivation for keeping the number low.
\end{itemize}

\begin{figure}
    \centering
    \includegraphics[width=\textwidth]{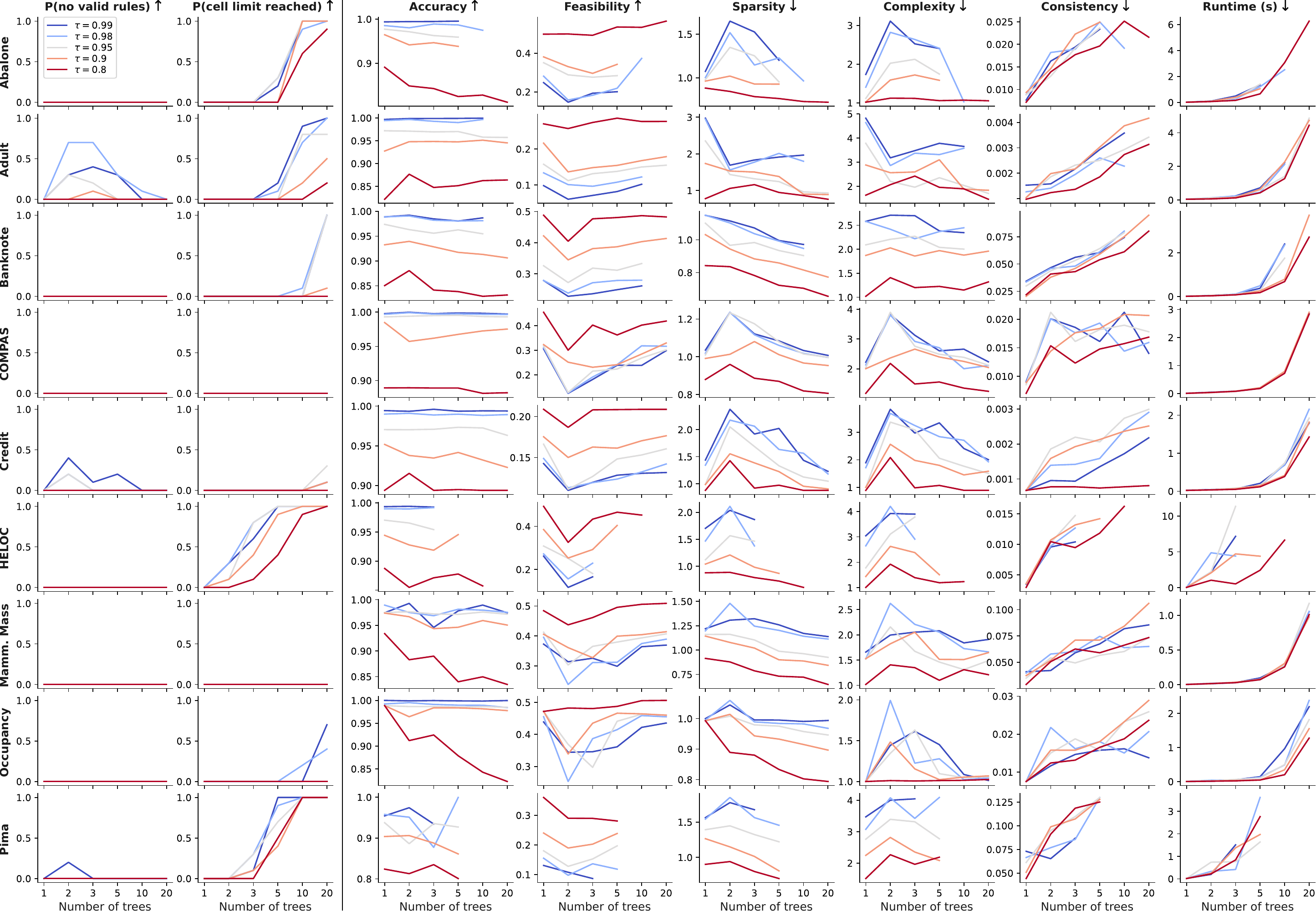}
    \vspace{-0.7cm}
    \caption{Hyperparameter study for all nine binary classification datasets.}
    \label{fig:hyperparam_full}
\end{figure}

\newpage
For the Adult dataset, we expand the hyperparameter study to consider variations in the feasibility threshold $\rho$ ($\in\{0.01,0.02,0.03,0.04,0.05\}$), and plot the results as a grid of heatmaps in Figure~\ref{fig:hyperparam_adult_multirho}. The probability of the two failure modes (no valid rules and cell limit reached) increase and decrease with $\rho$ respectively, both of which make sense because higher $\rho$ values yield fewer, larger rules. Overall, the fewest failures occur with $\rho=0.02$. Together with our informal finding that this value yields good results in practice, this explains our decision to use $\rho=0.02$ in all other experiments.

The consistency metric, as well as the algorithm's runtime, tend to decrease (i.e. improve) as $\rho$ is increased. The variations in other metrics exhibit less significant trends.

\begin{figure}[H]
    \centering
    \includegraphics[width=\textwidth]{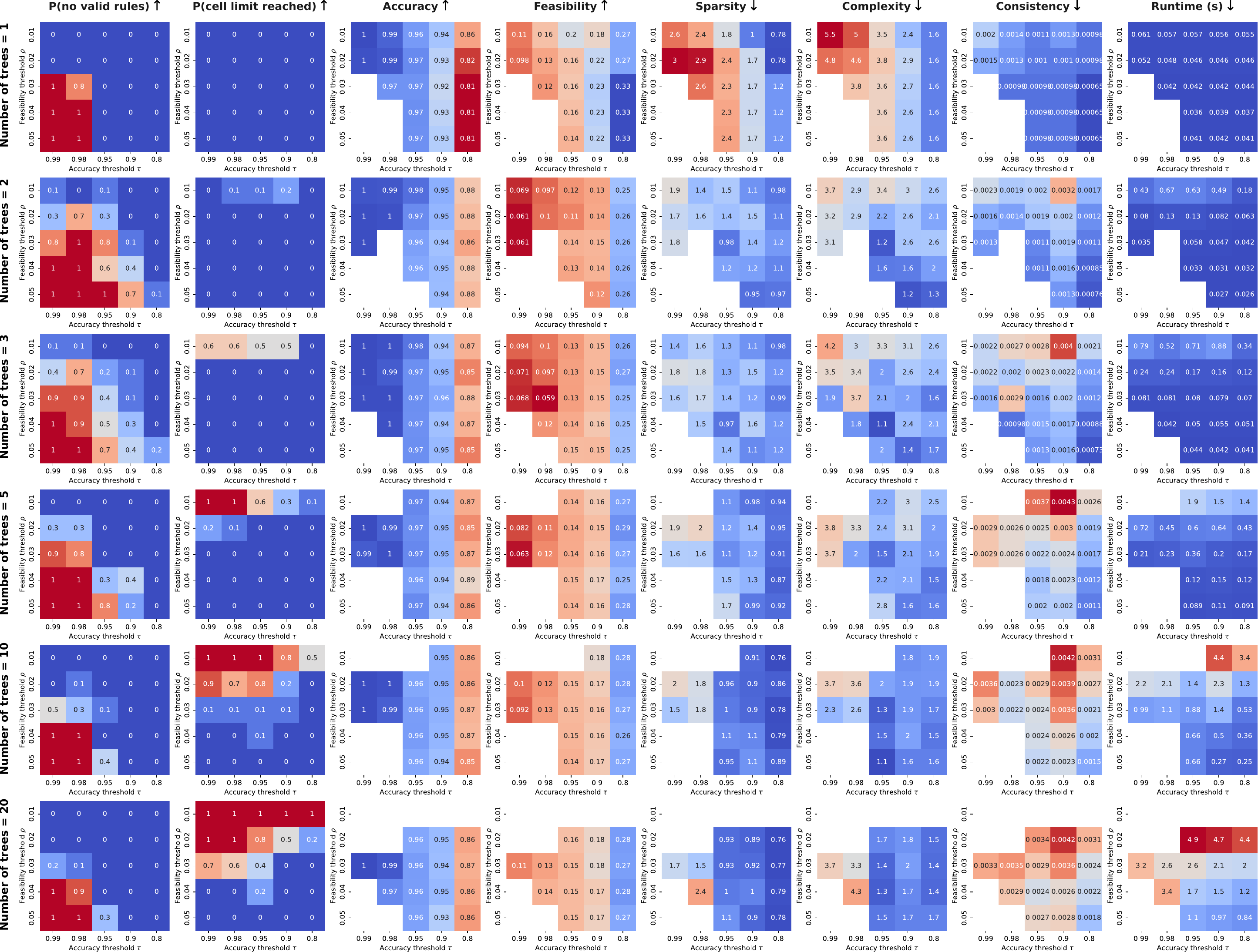}
    \vspace{-0.7cm}
    \caption{Extended hyperparameter study for the Adult data, also considering variable $\rho$.}
    \label{fig:hyperparam_adult_multirho}
\end{figure}

\newpage
\section{Baseline Details} \label{app:baselines}

For all three baselines, we use existing public Python implementations and retain default hyperparameter values (or values used in provided exemplar scripts or notebooks).

\subsection{\ares{}}

We use a third-party implementation by \citet{Kanamori2022counterfactual} at \url{https://github.com/kelicht/cet}. We retain most of the default hyperparameter values (\texttt{max\_rule=8, max\_rule\_length=8, discretization\_bins=10, max\_candidates=50}), but change \texttt{minimum\_support} from \texttt{0.6} to \texttt{0.05}, following \citet{Kanamori2022counterfactual} themselves (see \texttt{userstudy.py} on the GitHub repository). The implementation includes a \texttt{tuning} method to automatically set the objective weighting hyperparameters $\lambda$. We use all default hyperparameter values for this method (\texttt{max\_change\_num=4, cost\_type=TLPS, gamma=1.0, lambda\_acc=[1.0], lambda\_cov=[1.0], lambda\_cst=[1.0], objective=origin}).

A slight complication is that the implementation can only natively produce counterfactuals for cases where $f(x^0)=1$. To produce counterfactuals for $f(x^0)=0$, we artificially invert the model outputs and run the algorithm a second time.

\subsection{\lore{}}

We use the original authors' implementation at \url{https://github.com/kdd-lab/XAI-Lib}, which they recommend over an older version at \url{https://github.com/riccotti/LORE}. We use the same hyperparameter values for both the \texttt{fit} method (\texttt{neigh\_type=rndgen, size=1000, ocr=0.1, ngen=10}) and the \texttt{explain} method (\texttt{samples=1000, use\_weights=True, metric=neuclidean}) as in \texttt{tabular\_explanations\_example.ipynb} on the \texttt{XAI-Lib} GitHub repository.

In general, \lore{} returns a list of counterfactual rules. To enable direct comparison with the other methods, we retain only the single rule with the fewest finite boundaries (lowest complexity) with ties broken by taking the earliest in the returned list. In rare cases, \lore{} can also fail to return any rules. We exclude such cases from our evaluation.

The implementation handles categorical features using one-hot encoding but proceeds to treat each category feature as if it were numerical, allowing arbitrary rule boundaries that may fall outside the meaningful $[0, 1)$ range. It is unclear whether the authors intended this, but it is visible in \texttt{tabular\_explanations\_example.ipynb} on the \texttt{XAI-Lib} repository. We apply a post hoc correction to category feature boundaries, snapping those inside the $[0, 1)$ range to $0.5$, and ignoring those outside the range. The authors' na\"ive treatment of categoricals can also yield impossible `multi-hot' rules where more than one category feature is positive. In such cases, we apply another correction to choose one of these categories at random.

\subsection{\rfocse{}}

We use the original author's implementation at \url{https://github.com/rrunix/libfastcrf}, using default hyperparameter values for the \texttt{batch\_extraction} function (\texttt{max\_distance=-1, search\_closest=True, max\_iterations=-1, epsilon=0.0005}).

The authors present \rfocse{} as a method for generating counterfactual rules for random forests given direct access to their internal rule structure, but in this work, we are interested in explaining arbitrary black box models.
To adapt \rfocse{} for this context, we first fit a random forest surrogate to a training set labelled with model outputs, then apply the algorithm to the surrogate (i.e. mirroring our own approach). We use a \texttt{scikit-learn} \texttt{RandomForestClassifier} with \texttt{n\_estimators=10} and otherwise default hyperparameters, which matches the models studied in the \rfocse{} paper \cite{fernandez2020random}.

In that paper, it is stated that \textit{``categorical and binary variables are represented using numerical encoding''}, and rules are constructed as if those features were numerical. Reviewing the GitHub repository (specifically \texttt{research\_paper/dataset\_reader.py}), it appears that the numerical encoding is based on an alphabetical ordering of category names, and the rule examples given in Table 6 of \cite{fernandez2020random} are consistent with this. This is clearly an arbitrary and restricting choice, but we retain it to remain faithful to the original implementation.

\section{Experiment with Neural Network Model} \label{sec:app_mlp}

For the Adult dataset, we repeat the comparative evaluation in Figure~\ref{fig:main_results} for a neural network target model (specifically an \texttt{sklearn.neural\_network.MLPClassifier} with two hidden layers of $100$ units each and ReLU activations). The results are shown in Figure~\ref{fig:mlp_adult}, with the equivalent XGBoost results (from the main paper) copied below for reference. Although there are some changes in the rankings of methods on some of the desiderata, overall magnitudes are broadly similar in all cases. This indicates that neither our method, nor any of the baselines, exhibits any special sensitivity to the class of the target model.

\begin{figure}[H]
    \centering
    \includegraphics[width=\textwidth]{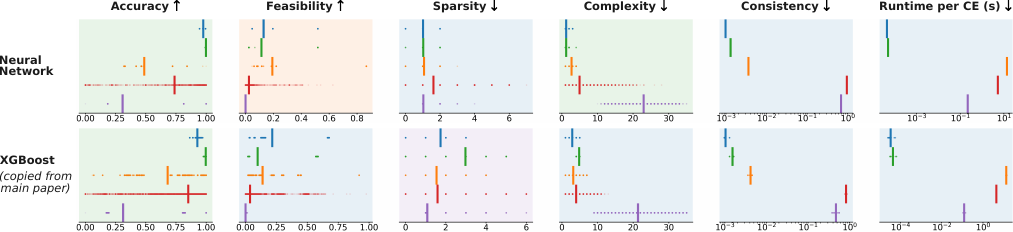}
    \vspace{-0.7cm}
    \caption{Performance of \ourstaulo{}, \ourstauhi{}, \ares{}, \lore{} and \rfocse{} for neural network and XGBoost models (Adult).}
    \label{fig:mlp_adult}
\end{figure}






\end{document}